\documentclass[preprints,article,accept,moreauthors,pdftex]{Definitions/mdpi}

\firstpage{1} 
\pubvolume{1}
\issuenum{1}
\articlenumber{0}
\pubyear{2021}
\copyrightyear{2020}
\datereceived{} 
\dateaccepted{} 
\datepublished{} 
\hreflink{https://doi.org/} 

\Title{scikit-dimension: a Python package for intrinsic dimension estimation}

\TitleCitation{scikit-dimension: a Python package for intrinsic dimension estimation}

\Author{Jonathan Bac $^{1,2,3}$*, Evgeny M. Mirkes $^{4,5}$\orcidA{}, Alexander N. Gorban $^{4,5}$\orcidG{}, Ivan Tyukin $^{4,5}$\orcidB{} and Andrei Zinovyev$^{1,2,3}$*\orcidC{} }

\address{%
$^{1}$ \quad Institut Curie, PSL Research University, Paris, France \\
$^{2}$ \quad INSERM, U900, Paris,France\\
$^{3}$ \quad CBIO-Centre for Computational Biology, Mines ParisTech, PSL Research University, Paris, France\\
$^{4}$ \quad Department of Mathematics, University of Leicester, Leicester, UK\\
$^{5}$ \quad Lobachevsky University, Nizhniy Novgorod, Russia
}

\corres{Correspondence: jonathan.bac@cri-paris.org (J,B.); andrei.zinovyev@curie.fr (A.Z.); Tel.: +33-156-246-989  (A.Z.)}

\abstract{Dealing with uncertainty in applications of machine learning to real-life data critically depends on the knowledge of intrinsic dimensionality (ID). A number of methods have been suggested for the purpose of estimating ID, but no standard package to easily apply them one by one or all at once has been implemented in Python. This technical note introduces \texttt{scikit-dimension}, an open-source Python package for intrinsic dimension estimation. \texttt{scikit-dimension} package provides a uniform implementation of most of the known ID estimators based on scikit-learn application programming interface to evaluate global and local intrinsic dimension, as well as generators of synthetic toy and benchmark datasets widespread in the literature. The package is developed with tools assessing the code quality, coverage, unit testing and continuous integration. We briefly describe the package and demonstrate its use in a large-scale (more than 500 datasets) benchmarking of methods for ID estimation in real-life and synthetic data. The source code is available at \url{https://github.com/j-bac/scikit-dimension}, the documentation is available at \url{https://scikit-dimension.readthedocs.io/}.}


\begin{document}

\section{Introduction}

We present \texttt{scikit-dimension}, an open-source Python package for global and local intrinsic dimension (ID) estimation. The package has two main objectives: (i) foster research in ID estimation by providing code to benchmark algorithms and a platform to share algorithms; (ii) democratize the use of ID estimation by provigin user-friendly implementations of  algorithms using Scikit-Learn application programming interface (API) \citep{pedregosa2011scikit}.

ID intuitively refers to the minimum number of parameters required to represent a dataset with satisfactory accuracy. The meaning of ``accuracy" can be different among various approaches. ID can be more precisely defined to be $n$ if the data lies closely to a $n$-dimensional manifold embedded in $R^d$ with little information loss, which corresponds to the so called ``manifold hypothesis" \citep{bishop1995neural,Fuku1982}. ID can be, however, defined without assuming the existence of a data manifold. In this case, data point cloud characteristics (e.g., linear separability or pattern of covariance) are compared to a model $n$-dimensional distribution (e.g., uniformly sampled $n$-sphere or $n$-dimensional isotropic Gaussian distribution) and the term ``effective dimensionality" is sometimes used as such $n$ which gives most similar characteristics to the one measured in the studied point cloud \citep{albergante2019estimating,del2020effective}. In \texttt{scikit-dimension}, these two notions are not distinguished. 

The knowledge of ID is important to determine the choice of machine learning algorithm and anticipate the uncertainty of its predictions. The well-known \textit{curse of dimensionality}, which states that many problems become exponentially difficult in high  dimensions, does not depend on the number of features but on the dataset’s ID \citep{jiang2018trust}. More precisely, the effects of dimensionality curse are expected to be manifested when $ID \gg ln(M)$, where $M$ is the number of data points \citep{Bac2020,hino2017ider}.

Current ID estimators have diverse operating principles (we refer the reader to \citep{Campadelli2015} for an overview). Each ID estimator is developed based on a selected feature (such as number of data points in a sphere of fixed radius, linear separability or expected normalized distance to the closest neighbour) which scales with $n$: therefore, various ID estimation methods provide different ID values. Each dataset can be characterized by a unique \textit{dimensionality profile} of ID estimations according to different existing methods, which can serve as an important signature for choosing the most appropriate data analysis method.

Dimensionality estimators that provide a single ID value for the whole dataset belong to the category of global estimators. However, datasets can have complex organizations and contain regions with varying dimensionality \citep{Bac2020}. In such a case, they can be explored using local estimators, which estimate ID in local neighborhoods around each point. The neighbourhoods are typically defined by considering the $k$ closest neighbours. Such approaches also allow repurposing global estimators as local estimators. 

The idea behind local ID estimation is to operate at a scale where the data manifold can be approximated by its tangent space \citep{Camastra2016}. In practice, ID is sensitive to scale and choosing neighbourhood size is a trade-off between opposite requirements \citep{Little,Campadelli2015}: ideally, the neighbourhood should be big relative to the scale of the noise, and contain enough points. At the same time, it should be small enough to be well approximated by a flat and uniform tangent space.

We perform benchmarking of 19 ID estimators on a large collection of real-life and synthetic datasets. Previously, estimators were benchmarked based mainly on artificial datasets representing uniformly sampled manifolds with known ID \cite{albergante2019estimating,Hein,Campadelli2015}, comparing them for the ability to estimate the ID value correctly. Several ID estimators were used on real-life datasets to evaluate the degree of dimensionality curse in a study of various metrics in data space\cite{mirkes2020entropy}. Here we benchmark ID estimation methods focusing on their applicability to a wide range of datasets of different origin, configuration and size. We also look at how different ID estimations are correlated, and show how \texttt{scikit-dimension} can be used to derive a consensus measure of data dimensionality, by averaging multiple individual measures. The latter can be a robust measure of data dimensionality in various applications.

\texttt{scikit-dimension} was applied in several recent studies for estimating intrinsic dimensionality of real-life datasets \citep{Golovenkin2020,Zinovyev2021CellCycleModeling}.
 
\section{Materials and Methods}

\subsection{Software features}

\texttt{scikit-dimension} consists of two modules. The \textit{id} module provides ID estimators and the \textit{datasets} module provides synthetic benchmark datasets.

\subsubsection{\textit{id} module}
The \textit{id} module contains estimators based on:
\begin{itemize}
\item correlation (fractal) dimension (id.CorrInt) \citep{Grassberger1983}
\item manifold-adaptive fractal dimension (id.MADA) \citep{farahmand2007manifold}
\item method of moments (id.MOM) \citep{amsaleg2018extreme}
\item principal component analysis (id.lPCA) \citep{jackson1993stopping,Fukunaga1971,Fuku1982,Fan}
\item maximum likelihood (id.MLE)
\citep{hill1975simple,Levina2004,haro2008translated}
\item minimum spanning trees (id.KNN) \citep{Carter2010a}
\item estimators based on concentration of measure (id.MiND\_ML, id.DANCo, id.ESS, id.TwoNN, id.FisherS, id.TLE) \citep{rozza2012novel,DANCo,Kerstin,Facco2017a,albergante2019estimating,GORBAN2018303,amsaleg2019intrinsic}.
\end{itemize}

The description of the method principles is provided together with the package documentation at \url{https://scikit-dimension.readthedocs.io/} and in reviews \citep{Hein,Bac2020,del2020effective}. 

\subsubsection{\textit{datasets} module}
The \textit{datasets} module allows user to test estimators on synthetic datasets. It can generate several low-dimensional toy datasets to play with different estimators as well as a set of synthetic manifolds commonly used to benchmark ID estimators, introduced by \citep{Hein} and further extended in \citep{rozza2012novel,Campadelli2015}.

\end{paracol}

\begin{figure}[H]
\widefigure
\includegraphics[width=16cm]{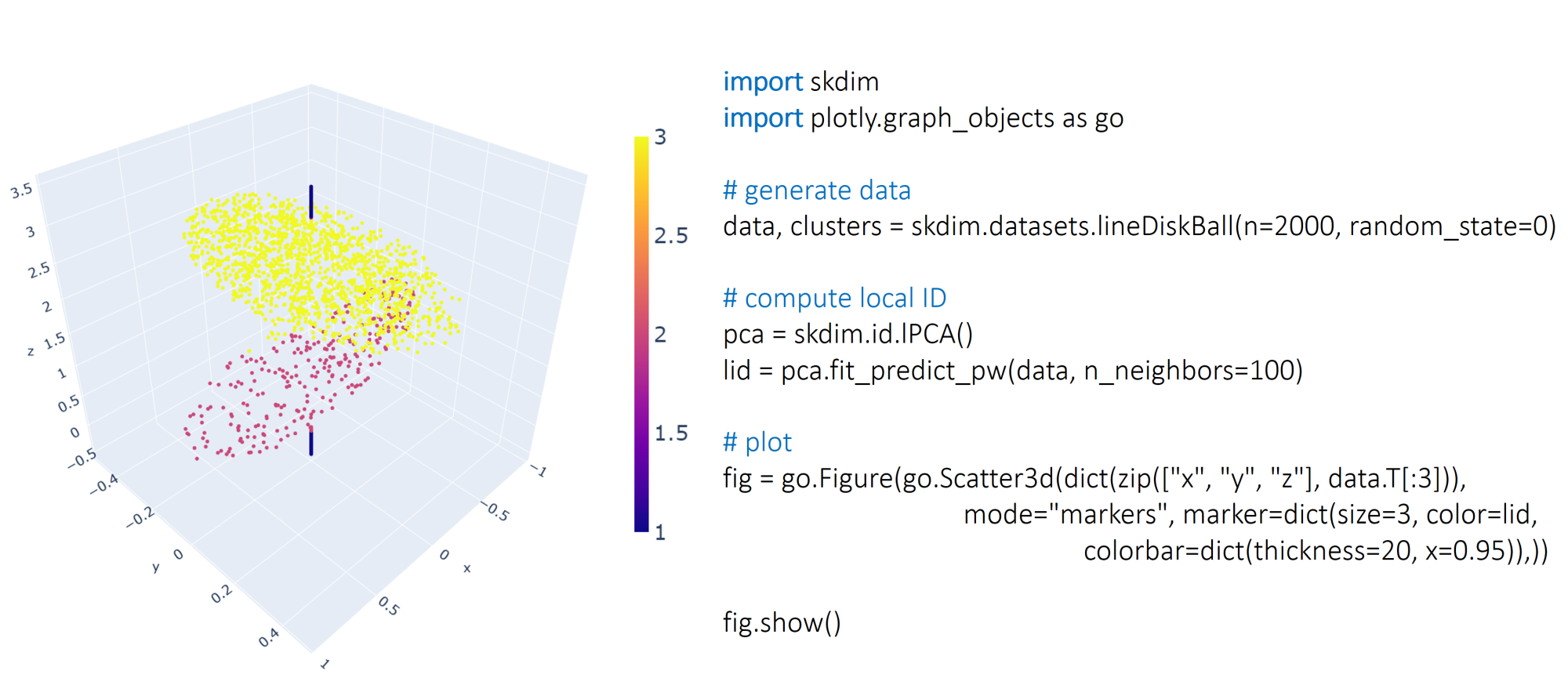}
\caption{Example usage: generating the Line-Disk-Ball dataset \citep{hino2017ider}) which has clusters of varying local ID, and coloring points by estimates of local ID obtained by id.lPCA. }
\end{figure}
\begin{figure}[H]	
\widefigure
\includegraphics[width=16cm]{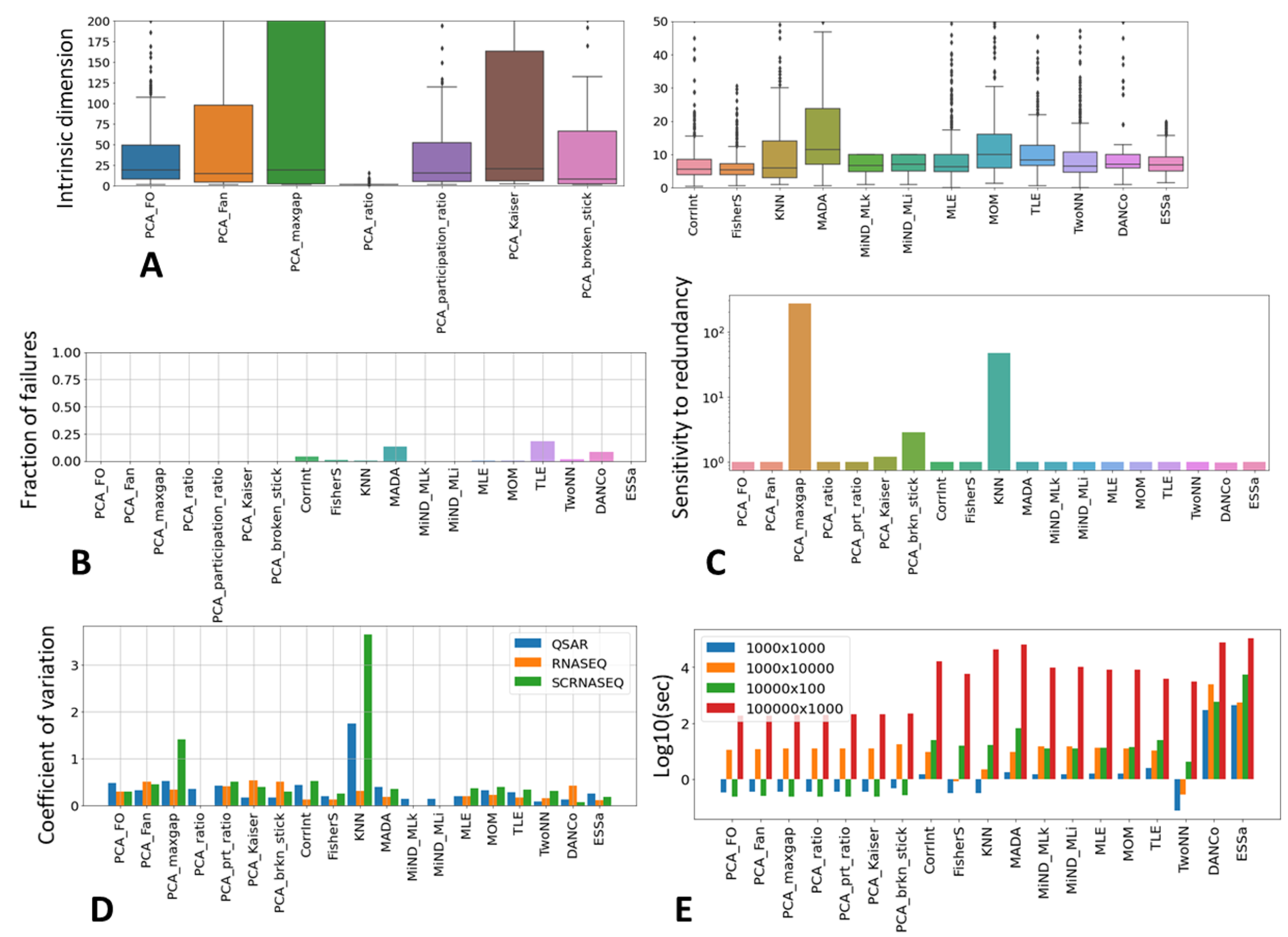}
\caption{Illustrating different ID method general characteristics: A) range of estimated ID values; B) ability to produce interpretable (positive finite value) result; C) sensitivity to feature redundancy (after duplicating matrix columns); D) uniform ID estimation across datasets of similar nature; E) computational time needed to compute ID for matrices of four characteristic sizes.\label{MethodPerformance}}
\end{figure}

\begin{paracol}{2}
\switchcolumn

\subsection{Development}

\texttt{scikit-dimension} is built according to the scikit-learn API \citep{pedregosa2011scikit} with support for Linux, MacOS, and Windows and Python $>=$ 3.6. Code style and API design is based on the guidelines of scikit-learn, with NumPy \citep{harris2020array} documentation format, and continuous integration on all three platforms. The online documentation is built using  \hyperlink{https://www.sphinx-doc.org/}{Sphinx} and hosted with \hyperlink{https://readthedocs.org/}{ReadTheDocs}.

\subsection{Dependencies}

\texttt{scikit-dimension} depends on a limited number of external dependencies on the user side for ease of installation and maintenance:
\begin{itemize}
\item Matplotlib \citep{Hunter:2007} \item Pandas \citep{reback2020pandas}
\item Scikit-learn \citep{pedregosa2011scikit}
\item Numba \citep{lam2015numba}
\item SciPy \citep{2020SciPy-NMeth} \item NumPy \citep{harris2020array} 
\end{itemize}

\subsection{Related software}

Related open source software for ID estimation have previously been developed in different languages such as R, MATLAB or C++ and contributed to the development of \texttt{scikit-dimension}.
    
In particular, \citep{intrinsicDimension,hino2017ider,you2020rdimtools}, provide extensive collections of ID estimators and datasets for R users, with \citep{you2020rdimtools} additionally focusing on dimension reduction algorithms. Similar resources can be found for MATLAB users \citep{IntDim,IDLombardi,drtoolbox,TLE}. Benchmarking many of the methods for ID estimation included in this package was performed in \cite{mirkes2020entropy}. Finally there exist several packages implementing standalone algorithms; in particular for Python, we refer the reader to complementary implementations of the GeoMLE, full correlation dimension, GraphDistancesID algorithms \citep{geomle, pyFCI, graphID}.

To our knowledge, \texttt{scikit-dimension} is the first Python implementation of an extensive collection of ID methods. Compared to similar efforts in other languages, the package puts emphasis on estimators quantifying various properties of high-dimensional data geometry such as concentration of measure. It is the only package to include ID estimation based on linear separability of data, using Fisher discriminants \citep{albergante2019estimating, bac2020local, gorban2019unreasonable,GORBAN2018303}.

\section{Results}

\subsection{Benchmarking \texttt{scikit-dimension} on a large collection of datasets}

In order to demonstrate applicability of \texttt{scikit-dimension} to a wide range of real-life datasets of various configurations and sizes, we performed a large-scale benchmarking of \texttt{scikit-dimension} using the collection of datasets from \texttt{OpenML} repository \cite{OpenML2013}. We selected those datasets having at least 1000 observations and 10 features, without missing values. We excluded those datasets which were difficult to fetch either because of their size or an error in the \texttt{OpenML} API. After filtering out repetitive entries, 499 datasets were collected. The number of observations in them varied from 1002 to 9199930 and the number of features varied from 10 to 13196. We focused only on numerical variables, and we subsampled the number of rows in the matrix to maximum 100000. All dataset features were scaled to unit interval using Min/Max scaler in Python. In addition, we filtered out approximate non-unique columns and rows in the data matrices since some of the ID methods could be affected by the presence of identical (or approximately identical) rows or columns.

We added to the collection 18 datasets, containing single-cell transcriptomic measurements, from the CytoTRACE study\citep{CytoTRACE2000} and 4 largest datasets from The Cancer Genome Atlas (TCGA), containing bulk transcriptomic measurements. Therefore, our final collection contained 521 datasets altogether.

\subsubsection{\texttt{scikit-dimension} ID estimator method features}

We systematically applied 19 ID estimation methods from \texttt{scikit-dimension}, with default parameter values, including 7 methods based on application of principal component analysis ("linear" or PCA-based ID methods), and 12 based on application of various other principles, including correlation dimension and concentration of measure-based methods ("nonlinear" ID methods). 

For KNN and MADA methods we had to further subsample the data matrix to maximum 20000 rows, otherwise they were too greedy in terms of memory consumption. Moreover, DANCo and ESS methods appeared to be too slow, especially in the case of a large number of variables: therefore, we made ID estimations in these cases on small fragments of data matrices. Thus, for DANCo the maximum matrix size was set to 10000x100, and for ESS to 2000x20. The number of features was reduced for these methods when needed, by using PCA-derived coordinates, and the number of observations were reduced by random subsampling.

In the Table~\ref{tab1}, we provide the summary of characteristics of the tested methods. In more detail, the following method features have been evaluated (see Figure~\ref{MethodPerformance}). 

Firstly, we simply looked at the ranges of ID values produced by the methods across all the datasets. These ranges varied significantly between the methods, especially for the linear ones (Figure~\ref{MethodPerformance},A).

Secondly, we tested the methods with respect to their ability to successfully compute ID as a positive finite value. It appeared that certain methods (such as MADA and TLE), in a certain number of cases produced a significant fraction of uninterpretable estimates (such as ``nan" or negative value), Figure~\ref{MethodPerformance},B. We assume that in most of such cases, the problem with ID estimation is caused by the method implementation, not anticipating certain relatively rare data point configurations, rather than the methodology itself, and a reasonable ID estimate always exists. Therefore, in case of a method implementation returning uninterpretable value, for further analysis, we considered it possible to impute the ID value from the results of application of other methods, see below.

\end{paracol}
\begin{table}[H] 
\small
\caption{Summary table of ID methods characteristics. The qualitative score changes from "- - -" (worst) to "+++" (best).\label{tab1}}
\begin{tabularx}{\textwidth}{XXXXXXXX}
\toprule
\textbf{Method name}& \textbf{Short \newline name(s)}	&\textbf{Ref(s)}& \textbf{Valid \newline result} & \textbf{Insensitivity \newline to redundancy} & \textbf{Uniform \newline ID estimate \newline in similar \newline datasets} & \textbf{Performance \newline with many \newline observations} & \textbf{Performance \newline with many \newline features} \\
\midrule
PCA Fukunaga-Olsen&PCA FO, PFO&\citep{Fukunaga1971,mirkes2020entropy}&+++&+++&+++&+++&+++\\
PCA Fan&PFN&\citep{Fan}&+++&+++&+++&+++&+++\\
PCA maxgap&PMG&\citep{Johnsson2015}&+++&- - -&+&+++&+++\\
PCA ratio&PRT&\citep{jolliffe2002principal}&+++&+++&+&+++&+++\\
PCA participation ratio&PPR&\citep{jolliffe2002principal}&+++&+++&++&+++&+++\\
PCA Kaiser&PKS&\citep{Kaiser1960TheAO,Giuliani2017}&+++&-&+++&+++&+++\\
PCA broken stick&PBS&\citep{FRONTIER197667,Cangelosi2007}&+++&- -&+++&+++&+++\\
Correlation (fractal) dimensionality&CorrInt, CID&\citep{Grassberger1983}&+&+++&++&+&+\\
Fisher separability&FisherS, FSH&\citep{GORBAN2018303,albergante2019estimating}&++&+++&+++&++&+++\\
K-nearest neighbours&KNN&\citep{Carter2010a}&++&- -&- -&-&++\\
Manifold-adaptive fractal dimension&MADA, MDA&\citep{farahmand2007manifold}&-&+++&+++&-&+\\
Minimum neighbor distance—ML&MIND\_ML,\newline MMk, \newline  MMi&\citep{rozza2012novel}&+++&+++&++&++&+\\
Maximum likelihood&MLE&\citep{Levina2004}&++&+++&++&++&+\\
Methods of moments&MOM&\citep{amsaleg2018extreme}&+++&+++&+++&++&+\\
Estimation within tight localities&TLE&\citep{amsaleg2019intrinsic}&- -&+++&+++&++&+\\
Minimal neighborhood information&TwoNN,\newline TNN&\citep{Facco2017a}&++&+++&+++&++&+++\\
Angle and norm concentration&DANCo,\newline DNC&\citep{DANCo}&+&+++&+++&- - -&- - -\\
Expected simplex skewness&ESS&\citep{Johnsson2015}&+++&+++&+++&- - -&- - -\\
\bottomrule
\end{tabularx}
\end{table}
\begin{paracol}{2}
\switchcolumn

Thirdly, for a small number of datasets we performed a test of their sensitivity to the presence of strongly redundant features. For this purpose, we duplicated all features in a matrix and recomputed ID. The resulting sensitivity is the ratio between the ID computed for the larger matrix and the ID computed for the initial matrix, having no duplicated columns. It appears that despite most of the methods being robust with respect to such matrix duplication, some (such as PCA-based broken stick or the famous Kaiser methods popular in various fields such as biology \citep{Giuliani2017,Cangelosi2007}) tend to be very sensitive, Figure~\ref{MethodPerformance},C, which is compliant with some previous reports \citep{mirkes2020entropy}.

Some of the datasets in our collection could be combined in homogeneous groups according to their origin, such as the data coming from Quantitative structure–activity relationship (QSAR)-based quantification of a set of chemicals. The size of the QSAR fingerprint for the molecules is the same in all such datasets (1024 features): therefore, we could assume that the estimate of ID should not vary too much across the datasets from the same group. We computed the coefficient of variation of ID estimate across three such dataset groups, which revealed that certain methods tend to provide less stable estimations than the others, Figure~\ref{MethodPerformance},D.

Finally, we recorded the computational time needed for each method. We found that the computational time could be estimated with good precision ($R^2>0.93$ for all ID estimators) using multiplicative model: $Time = c\times N_{obj}^\alpha \times N_{var}^\beta$, where $N_{obj}$ and $N_{var}$ are number of objects and features in a dataset, correspondingly. Using this model fit for each method, we estimated the time needed to estimate ID for data matrices of four characteristic sizes, Figure~\ref{MethodPerformance},E. 

\end{paracol}
\begin{paracol}{2}
\switchcolumn

\end{paracol}
\begin{figure}[H]
\widefigure
\includegraphics[width=16cm]{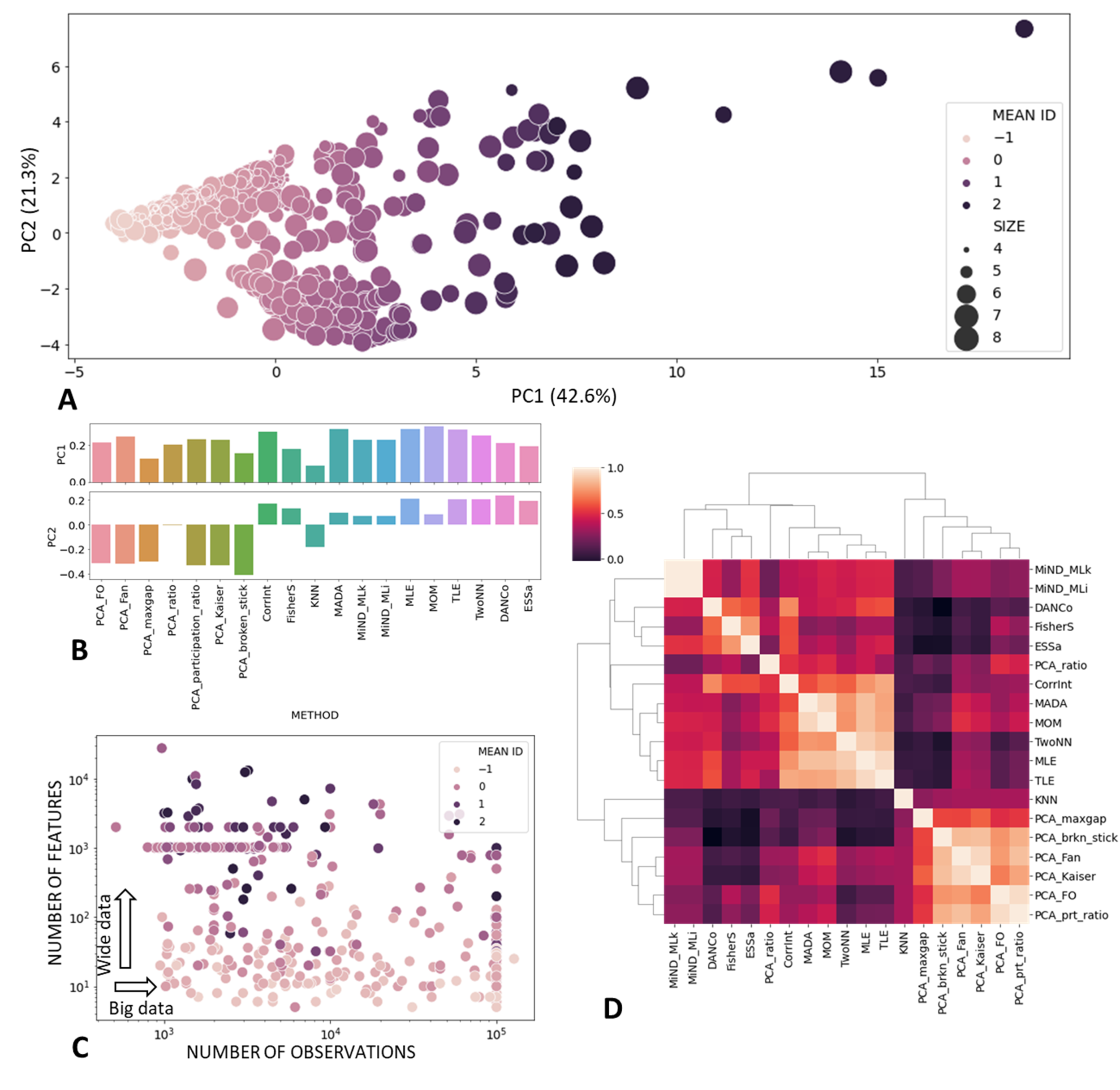}
\caption{Characterizing \texttt{OpenML} dataset collection in terms of ID estimates. A) PCA visualizations of datasets characterized by vectors of 19 ID measures. Size of the point corresponds to the logarithm of the number of matrix entries ($N_{obj}\times N_{var}$). The color corresponds to the mean ID estimate taken as the mean of all ID measure z-scores. B) Loadings of various methods into the first and the second principal component from A). C) Visualization of the mean ID score as a function of data matrix shape. The color is the same as in A). D) Correlation matrix between different ID estimates computed over all analyzed datasets. \label{MethodMetaanalysis}}
\end{figure}
\begin{paracol}{2}
\switchcolumn

\subsubsection{\texttt{scikit-dimension} ID estimates metanalysis}

After application of \texttt{scikit-dimension}, each dataset was characterized by a vector of 19 measurements of intrinsic dimensionality. The resulting matrix of ID values contained 2.5\% missing values which were imputed using the standard IterativeImputer from \texttt{sklearn} Python package. 

\newpage

\end{paracol}
\begin{figure}[H]
\widefigure
\includegraphics[width=15cm]{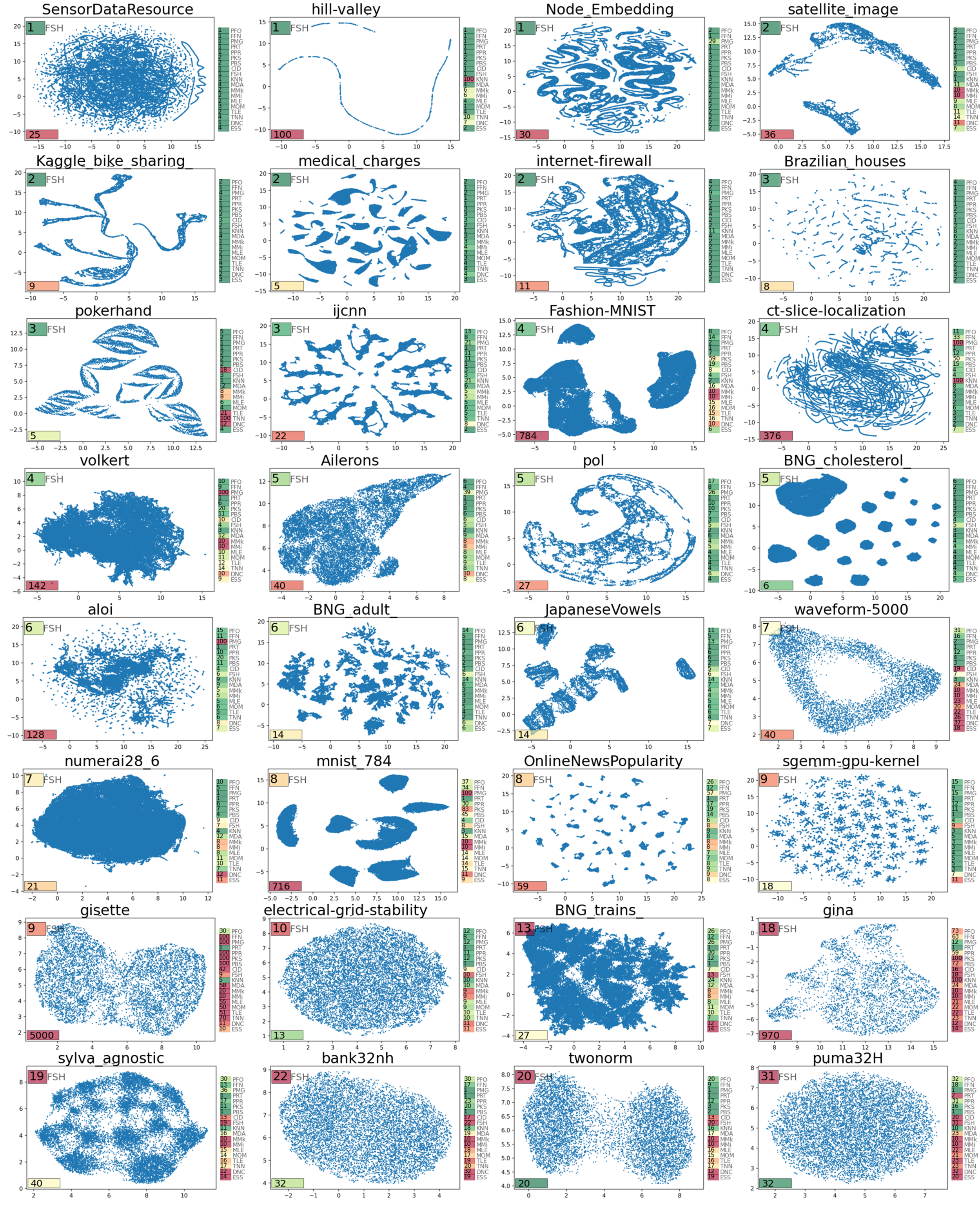}
\caption{A gallery of UMAP plots computed for a selection of datasets from \texttt{OpenML} collection, with indication of ID estimates, ranked by the ID value estimated using Fisher separability-based method (indicated in the left top corner). The ambient dimension of the data (number of features $N_{var}$) is indicated in the bottom left corner, and the color reflects the $ID/N_{var}$ ratio, from red (close to 0.0 value) to green (close to 1.0). On the right from the UMAP plot, all 19 ID measures are indicated, with color mapped to the value range, from green (small dimension) to red (high dimension). \label{UMAP}}
\end{figure}
\begin{paracol}{2}
\switchcolumn

Using the imputed matrix and scaling it to z-scores, we performed principal component analysis (Figure~\ref{MethodMetaanalysis},A,B). The first principal component explained 42.6\% percent of the total variance in ID estimations, with all of the methods having positive and comparable loadings to the first principal component. This justifies the computation of the ``consensus" intrinsic dimension measure, which we define here as the mean value of individual ID estimate z-scores. Therefore, the mean ID can take negative or positive values, roughly dividing the datasets into ``lower-dimensional" and ``higher-dimensional" (Figure~\ref{MethodMetaanalysis},A,C). The consensus ID estimate weakly negatively correlated with  with the number of observations (Pearson $\rho=-0.25$, p-value=$10^{-9}$) and positively correlated with the number of features in the dataset (r=0.44, p-value=$10^{-25}$). Nevertheless, even for the datasets with similar matrix shapes, the mean ID estimate could be quite different (Figure~\ref{MethodMetaanalysis},C). 

The second principal component explained 21.3\% of the total variance in ID estimates. The loadings of this component roughly differentiated between PCA-based ID estimates and ``non-linear" ID estimation methods, with one exception in the case of the KNN method. 

We computed the correlation matrix between the results of application of different ID methods (Figure~\ref{MethodMetaanalysis},D), which also distinguished two large groups of PCA-based and ``non-linear" methods. Furthemore, non-linear methods were split into the group of methods producing results similar to correlation (fractal) dimension (CorrInt, MADA, MOM, TwoNN, MLE, TLE) and methods based on concentration of measure phenomena (FisherS, ESS, DANCo, MiND\_ML). 

In order to illustrate the relation between the dataset geometry and the intrinsic dimension, we produced a gallery of Uniform Manifold Approximation and Projection (UMAP) dataset visualizations, with indication of the ambient dataset dimension (number of features) and the estimated ID using all methods, Figure~\ref{UMAP}. One of the conclusions that can be made from this analysis is that UMAP visualization is not insightfull for truely high-dimensional datasets (starting from ID=10 estimated by FisherS method). Also, some datasets having large ambient dimensions, were characterized by low ID, by most of the methods (e.g., 'hill-valley' dataset).

\section{Conclusions}

\texttt{scikit-dimension} is the first to our knowledge package implemented in Python, containing implementations of most used estimators of data intrinsic dimensionality. 

Benchmarking \texttt{scikit-dimension} on a large collection of real-life and synthetic datasets revealed that different estimators of intrinsic dimensionality possess internal consistency and that the ensemble of ID estimators allows us to achieve more robust classification of datasets into low- or high-dimensional.

Future releases of \texttt{scikit-dimension} will continuously seek to incorporate new estimators and benchmark datasets introduced in the literature, or new features such as alternative nearest neighbor search for local ID estimates. The package will also include new ID estimators which can be derived using most recent achievements in understanding the properties of high-dimensional data geometry\citep{Gorban2021HighDimSepar,GRECHUK202133}.

\vspace{6pt} 



\authorcontributions{Conceptualization, J.B., A.Z., I.T., A.N.G.; methodology, J.B., E.M., A.N.G.; software, J.B., E.M. and A.Z.; formal analysis, J.B., E.M., A.Z.; data curation, J.B. and A.Z.; writing---original draft preparation, J.B. and A.Z.; writing---review and editing, all authors; supervision, A.Z. and A.N.G. All authors have read and agreed to the published version of the manuscript.}

\funding{The work was supported by the Ministry of Science and Higher Education of the Russian Federation (Project No. 075-15-2021-634), by the French government under management of Agence Nationale de la Recherche as part of the ``Investissements d’Avenir" program, reference ANR-19-P3IA-0001 (PRAIRIE 3IA Institute), by the Association Science et Technologie, the Institut de Recherches Internationales Servier and the doctoral school Frontières de l’Innovation en Recherche et Education Programme Bettencourt. I.T. was supported by the UKRI Turing AI Acceleration Fellowship (EP/V025295/1).}



\dataavailability{The datasets used in this study were retrieved from public sources, namely \texttt{OpenML} repository, CytoTRACE web-site \url{https://cytotrace.stanford.edu/} (section "Downloads"), from the Data Portal of National Cancer Institute \url{https://portal.gdc.cancer.gov/}.} 


\conflictsofinterest{The authors declare no conflict of interest.} 





\end{paracol}

\reftitle{References}


\externalbibliography{yes}
\bibliography{skdim_preprint}

\end{document}